\pdfoutput=1

\documentclass[11pt]{article}

\usepackage{acl}

\usepackage{times}
\usepackage{latexsym}
\usepackage[export]{adjustbox}

\usepackage[T1]{fontenc}

\usepackage[utf8]{inputenc}

\definecolor{light-gray}{gray}{0.90}
\definecolor{dark-gray}{gray}{0.30}
\definecolor{ourlightblue}{HTML}{E0ECF7}
\definecolor{ourdarkblue}{HTML}{092E6B}
\definecolor{msgrblue}{HTML}{4889f4}
\definecolor{msgrgray}{HTML}{e1e1e7}
\definecolor{msgrpaleblue}{HTML}{a9c6f5}
\definecolor{palegreen}{HTML}{c0eeC3}
\definecolor{palepurple}{HTML}{e5d1f8}
\definecolor{paleorange}{HTML}{f9dbb1}
\definecolor{palered}{HTML}{f8aca7}
\definecolor{paleyellow}{HTML}{f8f1a7}
\definecolor{brightergreen}{HTML}{63ec99}

\definecolor{botc}{rgb}{0.458, 0.488, 0.978}

\usepackage{microtype}

\usepackage{adjustbox}
\usepackage{graphicx}
\usepackage{multirow}
\usepackage{booktabs}
\usepackage{todonotes}
\usepackage{float}
\usepackage{xstring}
\usepackage{enumitem}
\usepackage[bottom]{footmisc}

\definecolor{lightblue}{HTML}{E0ECF7}
\definecolor{darkblue}{HTML}{092E6B}

\title{Learning New Skills after Deployment:\\
{\em Improving open-domain internet-driven dialogue with human feedback}}
\author{Jing Xu\\
  Meta AI 
  \And
  Megan Ung\\
  Meta AI 
  \And
  Mojtaba Komeili\\
 Meta AI 
  \AND
  Kushal Arora \\
  Meta AI \& \\
  Mila / McGill University
  \And
  Y-Lan Boureau\\
  Meta AI 
  \And
  Jason Weston \\
  Meta AI 
}

\begin{document}
\maketitle
\begin{abstract}
Frozen models trained to mimic static datasets can never improve their performance. 
Models that can employ internet-retrieval for up-to-date information and obtain 
feedback from humans during deployment provide the promise of both 
adapting  to new information, and improving their performance.
In this work we study how to improve internet-driven conversational skills
in such a learning framework.
We collect deployment data, which we make publicly available, of human interactions, and collect various types of human feedback -- including binary quality measurements, free-form text feedback, and fine-grained reasons for failure. We then study various algorithms for improving from such feedback, including standard supervised learning, rejection sampling, model-guiding and reward-based learning, in order to make recommendations on which type of feedback and algorithms work best. 
We find the recently introduced {\sc Director} model \cite{arora2022director}
shows significant improvements over other existing approaches. \end{abstract}

\section{Introduction}

Large language models employed as dialogue agents are primarily trained on human-written documents and human-human conversations collected from the web for pre-training \cite{conneau2019unsupervised,baumgartner2020pushshift},
and human-human crowdsourced conversations \cite{smith2020bst} for fine-tuning. The models are then used at inference time to conduct conversations with humans, with no further learning taking place 
\cite{adiwardana2020meena,roller2020recipes}.
Human-model conversations -- which are never seen at training time --
can  have a quite different distribution
to the original human-human training data used, and our current techniques can lose
performance due to lack of robustness to such deviations \cite{chollet2019measure,bengio2019system}.

\begin{figure}[t!]
  \centering
    \includegraphics[width=0.48\textwidth]{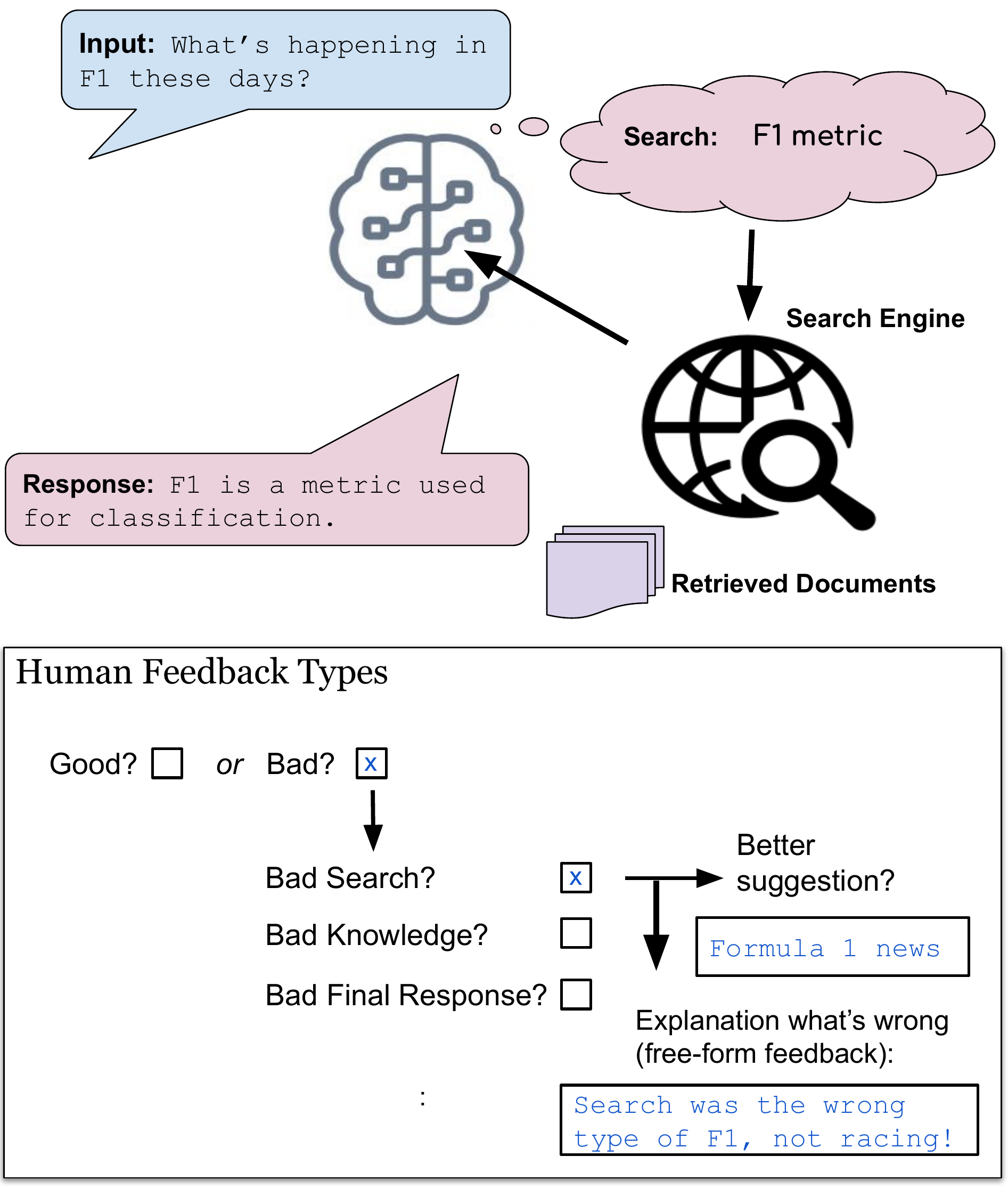}
  \caption{{\bf Using human feedback to improve open-domain internet-driven dialogue agents.} We compare various types of feedback (and corresponding  learning algorithms) in this work, such as binary feedback (good/bad), free-form text or supervised responses (better suggestions) for different modules of the system.
    \label{fig:diagram}
    }
\end{figure}

In this work, we study learning from the feedback collected during deployment of models in human-model conversations.
Such a setting has the opportunity to learn from within-distribution data, both in terms of the input contexts, but also the responses required (targets). Not only can this mean improvement in skills that are similar to the pre-train and fine-tune data, but potentially the learning of completely new skills -- that are desired by users of the system.
We thus take existing state of the art internet-augmented models such as BlenderBot 2 \cite{komeili2021internet,xu2021beyond} and SeeKeR \cite{shuster2022language}, deploy them to human crowdworkers, and experiment with various methods to learn from such interactions.
We thus first ask crowdworkers what topic and task they would like to talk about, in order to collect in-domain data, and then collect conversations involving these 
 skills. During the conversations we collect various kinds of human feedback, including binary feedback (good/bad), free-form conversational feedback, and the type of failure (search query-based, results-based, or final response-based), as well as suggestions for improvements (see \autoref{fig:diagram}).

We then explore a variety of methods for learning from feedback, and compare them in detailed experiments. In particular, we compare supervised learning methods, rejection sampling, model guiding and reward-based learning. Our findings are:
\begin{itemize}
    \item Taking advantage of modular feedback (feedback about particular errors from modules of the model, such as the search engine component) outperforms feedback about just the final response.
    \item Textual and binary feedback are also very useful signals, but not as much as modular feedback. 
    \item The recently introduced {\sc Director} method \cite{arora2022director}, when learning from binary feedback, works better than reranking or reward-based learning. 
    \item Combining multiple types of feedback, such as modular and binary feedback with {\sc Director}  provides the best results we obtained.
    \item Continual learning, whereby we retrain models on the feedback from previous rounds of deployment, improves results even further.  
    \item Despite collecting feedback from smaller (3B parameter) models, the data collection is useful for improving much larger (175B parameter) models.
\end{itemize}
We release the collected data and feedback, the models, and make the code publicly available for this work\footnote{\url{https://parl.ai/projects/fits}}.




\section{Related Work}

There are a number of existing methods for collecting human feedback from human-model conversations. Deployed models can be improved in symmetric conversations conducted between models and humans  during deployment by learning to mimic human conversationalists, as shown in the LIGHT dialogue game \cite{shuster2020deploying}. This is not directly applicable if the conversations are asymmetric, for example in the case of one speaker (human) who asks the questions, and the other (bot) who always answers, as there would be no human supervision of the answers. In the non-symmetric case, one can however try to make use of the textual response from humans when conversing with the bot, but alternative learning  methods must then be used. \citet{li2016learning} studies models that learn how to ask questions in order to learn from the answers, while \citet{li2016dialogue} learns from general textual feedback/comments, particularly in the case where the bot has produced a low quality response. Another approach is to learn a reward signal (positive or negative reaction) based on user textual responses, as shown in the ``self-feeding chatbot''  \cite{hancock2019learning}. Finally, rather than using conversational feedback, one can use sophisticated web-based UIs to collect data, for example stack ranking potential responses \cite{ouyang2022training,bai2022training}. 

Outside of the dialogue domain, there are numerous studies attempting to improve language skills from deployment, including  never-ending-learning from language data \cite{carlson2010toward}, learning for the web search task directly \cite{agichtein2006improving} or the Dynabench system which covers a number of NLP tasks \cite{kiela2021dynabench}.
\citet{nakano2021webgpt} also learns to use internet-augmentation for generation, like this work, but for question answering, not multi-turn dialogue.

\begin{table*}
\small
\center
\begin{tabular}{p{0.2\linewidth}|p{0.35\linewidth}|p{0.35\linewidth}}
Topic & Specific Task & Task Completion Description \\
\hline
Making healthy food & Find recipes on healthy foods & If the chatbot provided specific recipes on making healthy foods \\
\hline
I would like to learn about a type of pet &	I would like to learn about some hypoallergenic breeds of dogs, specifically, small dogs. &	If the chatbot could tell me some small dog breeds that are hypoallergenic, along with details about the breed's temperament, personality and any special requirements.\\
\hline
gravel driveway & choosing the correct gravel for your driveway & It would ask a variety of questions. It would ask the length of your driveway, the area you live in, and your price range. Then it would show you pictures of different types of gravel that fit the criteria. Lastly, it would provide broad steps on gravelling the driveway and estimated price ranges for project completion.\\
\hline
getting started with cycling &	what do I need to do to get started with road cycling	& The chatbot would tell me what kind of bicycle would be best for road cycling and the necessary accessories that a beginner needs.\\
\hline
Find child friendly places in a city &	Find child friendly resorts in Nassau Bahamas &	Pull up resorts in Nassau Bahamas, only show the resorts that are child friendly, give the star rating for each resort, show the child programs in the resort.\\
\end{tabular}
\caption{
A sample of the collected topics and task definitions. See 
\autoref{tab:dataset} for statistics on the overall dataset.
\label{tab:task_examples}
}
\end{table*}

\section{Deploying and Collecting Feedback}

\subsection{Open-domain internet-driven skills} \label{sec:task_defs}

To select an input distribution closely aligned with human preferences, we first collected a set of skills humans would like an AI powered text-messaging chatbot to possess. We instruct that
the hypothetical chatbot can talk about any topic, and has the ability to surf the internet for information.
We then asked each human annotator to provide:
\begin{itemize}
\item[(i)] a topic (1-10 words), 
\item[(ii)] three tasks related to the topic; and 
\item [(iii)] descriptions of how they would assess if the chatbot has completed those tasks.
\end{itemize}
See Appendix \autoref{sec:task_def_collection} for a screenshot of the task definition collection instructions, and further details.

Overall, we collected 1108 task types 
via 152 annotators, which cover diverse topics -- from making healthy food to loom weaving to Caribbean holidays.
Grouping them into types, they include question answering followed by discussion, providing ranked lists, providing reviews, summary generation, personal recommendations, reasoning/deductions (e.g., how to perform calculations), creativity (e.g., tell a joke), tutorials, instructions, and more. 
Many of these tasks require, or else are made simpler, by use of the internet, e.g., searching for particular entities or topics, and responding conditioned on pertinent results. 
Some examples are given in \autoref{tab:task_examples}.

\subsection{Conversing with models and receiving feedback} 

After collecting topic and task definitions, the next step is to  deploy conversational models (bots) that are asked to exhibit these skills. 
Human conversationalists select a task (out of two randomly chosen tasks) from the set collected in \autoref{sec:task_defs} and then ask the model to help them complete it over a series of conversational turns. The instructions emphasize that this should be a dialogue (``a back and forth conversation''), and hence the speakers should break up requests or information across messages so that it remains conversational.

\paragraph{Feedback types} \label{sec:feedback_types}
The human conversationalist is instructed
that the bot might not be perfect, in which case feedback can be given in order to improve the bot in the future. We collect various kinds of feedback, from lightweight feedback (binary label or free-form response) to detailed (multiple choice and fine-grained responses) such that in our experiments we can compare and contrast them in order to make recommendations on which kinds of feedback work best.

Hence after each dialogue turn we collect the following set of feedback types:
\begin{itemize}
    \item Binary feedback on whether the response was considered satisfactory or not.
    
    \item Free-form textual feedback on what was wrong in the case of an unsatisfactory response.
    \item Multi-choice input on how the bot could improve this turn:
    \begin{itemize}
        \item[(a)] using a better search query; or
        \item[(b)] paying more attention to relevant search results; 
        \item[(c)] some other issue; or
        \item[(d)] no issue (a good response).
    \end{itemize}
    \item In the case of selecting (a), the human is then asked what would be a more appropriate search query.
    \item In the case of (b), the human is presented the search results and asked to select a relevant portion.
    \item In the case of (c), the human is asked what would be an improved overall response.
\end{itemize}

\begin{table*}
\centering
\small
\begin{tabular}{lrrr|rrr|r}
  & \multicolumn{3}{c}{\bf v1} & \multicolumn{3}{c}{\bf v2} &  \\
{\bf Collected Data}  & Train & Valid & Test Seen  & Train &  Valid & Test Seen & Test Unseen \\
 \hline
 \hline
Number of Unique Tasks & 963 & 524 & 709 & 980 & 814 & 824 & 114 \\
Number of Dialogues & 5592 &  737 & 1230 & 9817 & 1848 & 1848 & 1221 \\
Number of Utterances & 77946 & 8490 & 19452 & 140702 & 22560 & 29860 & 17814 \\
Number of Bot Utterances & 38523 & 4245 & 9726 & 70351 & 11280 & 14930 & 8907 \\
Average Bot Utterances per Dialogue & 6.89 & 5.76 & 7.91 & 7.17 & 6.10 & 8.08 & 7.29\\
\hline
{\bf Feedback Breakdown} \\
Better Search Query & 5179 & 605 & 1167 & 8778 & 1425 & 1706 & 1036 \\
Better Results Usage & 6875 & 756 & 1527 & 11429 & 1796 & 2340 & 1310 \\
Better Response & 6601 & 714 & 1493 & 10812 & 1472 & 2382 & 1372\\
Good Response & 19868 & 2170 & 5539 & 39332 & 6587 & 8502 & 5189\\ 
Average Good Utterances per Dialogue & 3.55 & 2.94 & 4.50 & 4.01 & 3.56 & 4.60 & 4.25\\
\hline 
\end{tabular}
\caption{
Collected human-bot conversations and feedback data statistics for the dataset 
{\bf FITS} (Feedback for Interactive Talk \& Search), which we publicly release. Train v1 and v2 correspond to two rounds of continual learning, where the v1 data consists of conversations and feedback from our deployed base models, and the v2 data consists of conversations and feedback from deployed models trained using the v1 data.
\label{tab:dataset}
}
\end{table*}

\paragraph{Continuing the conversation} \label{sec:continuing_convo}

After feedback has been given, the conversation is continued. If multiple-choice option (a) was selected previously, the bot on this next turn is forced to use the ``gold'' search query given by the user. Similarly, for (b), the provided gold knowledge context is added to the input of the model. In the case of (c), the bot is simply bypassed, and it is assumed to have provided the given gold response. In this way, even for a poorly performing bot, headway can be made in the conversation towards completing the task, and collecting feedback on its subsequent stages. (Without such a procedure, the bot may just get stuck in a poor quality loop, and then there would be no choice but to abandon the conversation.)  

The conversation is continued until the human marks the task as complete or a maximum of 4 turns has been completed. 
When the task is complete we also collect a final rating (out of 5) for the bot's performance.

\subsection{Deployed Models} \label{sec:base_deployed}

We consider the following set of state of the art publicly available conversational models:
\begin{itemize}
    \item BlenderBot  (BB1) \cite{roller-etal-2021-recipes}; a 2.7B parameter Transformer model pre-trained and fine-tuned on dialogue data to exhibit conversational skills; however these models have no ability to use the internet, but simply generate responses given the dialogue context. 
    \item BlenderBot 2.0 (BB2) \cite{komeili2021internet,xu2021beyond}, 
    a 2.7B parameter model multi-tasked on the same tasks as BB1, and also with additional tasks which give it the ability to execute internet search queries and condition on the results using a fusion-in-decoder (FiD) \cite{izacard2020leveraging} style approach. The search query generator is a separate 400M parameter transformer.
    \item SeeKeR \cite{shuster2022language}; again uses a similar 2.7B parameter architecture, but utilizing the Knowledge-to-response (K2R) approach \cite{adolphs2021reason} which performs a multi-step generation procedure: first generating a relevant knowledge response, and then conditions on that to generate a final dialogue response. It is multi-tasked on the same training data as BB2, and in addition on some other knowledge-intensive tasks, such as QA tasks, as well. 
    \item OPT-175B \cite{zhang2022opt} and BB3-175B \cite{shuster2022bb3}: we compare the 175B language model OPT (either 0-shot or few-shot\footnote{Five examples are provided with input+response for each module in turn, mimicking the SeeKeR setup, see \citet{shuster2022bb3} for more details.}) with BlenderBot 3, which is fine-tuned with conversational datasets including modular supervision, and internet-augmentation, from our task.  
    This setting examines if our experiments and results are applicable to very large language models.
\end{itemize}

\subsection{Evaluation}

We can evaluate model performance during  conversations between humans and the deployed models, as humans are providing direct feedback on the conversational responses from the model. 
In particular we can measure the number of good responses (with no issue), the average final rating, 
and compute a  breakdown of error types (better search query, results or other issue).

\subsection{Collected Data} \label{sec:collected_data}

Overall, in our experiments we collect over 210k human-bot utterances
in over 14k dialogues (episodes), with feedback for each of the bot utterances. The data is split into three major portions:  v1, v2, and test unseen splits,
see \autoref{tab:dataset} for a full breakdown.

The {\bf v1 split}  consists of  dialogues 
conducted with one of our base deployed models (\autoref{sec:base_deployed}), and feedback was collected from those dialogues. We then split that data into train, valid and test dialogues. We use this data to then train several learning methods using the feedback from the v1 models. These new models are then  redeployed. 

The {\bf v2 split} consists of dialogues and feedback  with the new models that were trained using the v1 data. This data is again split into train, valid and test dialogues. We can then repeat this process and train models on the v2 data as well.

Finally, the {\bf unseen test} split consists of completely new skills (topics and tasks) unseen in the v1 and v2 splits, and is used to test transfer of v1 or v2 based models to these new skills.

\paragraph{Data Quality and Verification}
We also verified the quality of our data. For each conversation, we ask 3 human crowdworkers to rate the bot and human's performance and also assess if the bot was able to complete the given task. We consider the task as complete if 2 out of the 3 annotators labeled the task as complete. We see that in 90.4\% of the cases the task is completed. Note that with the feedback from the human (see \autoref{sec:continuing_convo}) the human-model conversation should always progress even if the model has errors so 
ideally if the human is doing a perfect job this would be 100\%.  We also assess the quality of the human conversationalist directly and ask annotators to ``\textit{rate the human's messages in defining, clarifying, and helping the bot complete the task on a scale from 1-5 (1 = was not helpful at all in helping the bot complete the task, 5 = guided the bot to complete the task).}'' For conversations where the task was completed, the human conversation partner's messages were rated at an average of 3.8. For conversations where the task was incomplete, the human conversation partner's messages were rated at an average of 3.5.

\section{Feedback Learning Methods}

In the following, we will describe the methods we will experiment with for learning from the collected human feedback.

\subsection{Supervised Learning of Responses} 

The easiest to use type of feedback, with perhaps the strongest learning signal, is a provided gold response by the user for a given dialogue context.
One can simply continue to fine-tune the model on the set of collected gold responses (from case (c) in \autoref{sec:feedback_types}). One can  optionally also add all the responses the bot itself made that were marked as good  to the fine-tune set as well (case (d) in \autoref{sec:feedback_types}). We use the validation set to choose the weighting between these two types of supervised data.

\subsection{Fine-grained Module supervision}

Using the multiple-choice feedback on the types of improvement, the model can learn to improve those individual components of the model. 
For BB2 and SeeKeR one can use provided gold search queries 
(case (a) in \autoref{sec:feedback_types})
directly to fine-tune the search query generation. 

Provided gold knowledge responses (relevant search results, case (b) in \autoref{sec:feedback_types})) are similarly easy to use for fine-tuning in the SeeKeR model because the model is already trained to generate such responses directly.
For BB2, there are no direct knowledge responses as this is implicit in FiD, so in that case we use a similar method to \citet{hancock2019learning} whereby we train in a supervised fashion with the knowledge response as a target, but add special tokens to both input and target to indicate this is not a standard dialogue response task. The goal is that this additional training signal  can then help learn useful features for the actual overall response task.

\subsection{Free-form Textual Feedback}

For free-form textual feedback, we can also use a similar approach 
and simply fine-tune with the feedback as targets, with special tokens appended to both the input context and the feedback target, again following \citet{hancock2019learning} which showed this approach can work.

\subsection{Rejection Sampling/Reranking} \label{sec:rerank}

Using the binary satisfaction feedback signal one can train a reward model. 
We employ a 311M parameter transformer pre-trained on  pushshift.io Reddit \cite{baumgartner2020pushshift} using a masked language model objective. Then, given the context and response concatenated as input, we  train it with a standard classification loss on our satisfaction task.
Such a model has multiple uses (see following subsections) but one obvious approach is to rerank generation candidates from the conversational  model using the reward model with the aim that the highest ranked provide the highest satisfaction. Such approaches have been employed in many use cases previously \cite{nie2020like,nakano2021webgpt,askell2021general,thoppilan2022lamda}. 

\subsection{Reward-based Learning}

Rejection sampling/reranking relies on the set of generated candidates containing at least one good candidate, and has no effect on the initial quality of the candidate generations themselves 
-- it only scores the final generated sequences.
We  next consider using a reward model trained via \autoref{sec:rerank} to train the generation model itself. Given training set contexts, we generate candidates, rerank the candidates, and select the highest ranking. We then train the generation model  to use those highest ranking candidates as targets, i.e. by fine-tuning with those targets. This is similar to the approach used in \citet{thoppilan2022lamda}.

\subsection{Model-guiding with {\sc Director}}

\begin{figure}[t!]
  \centering
    \includegraphics[width=0.5\textwidth]{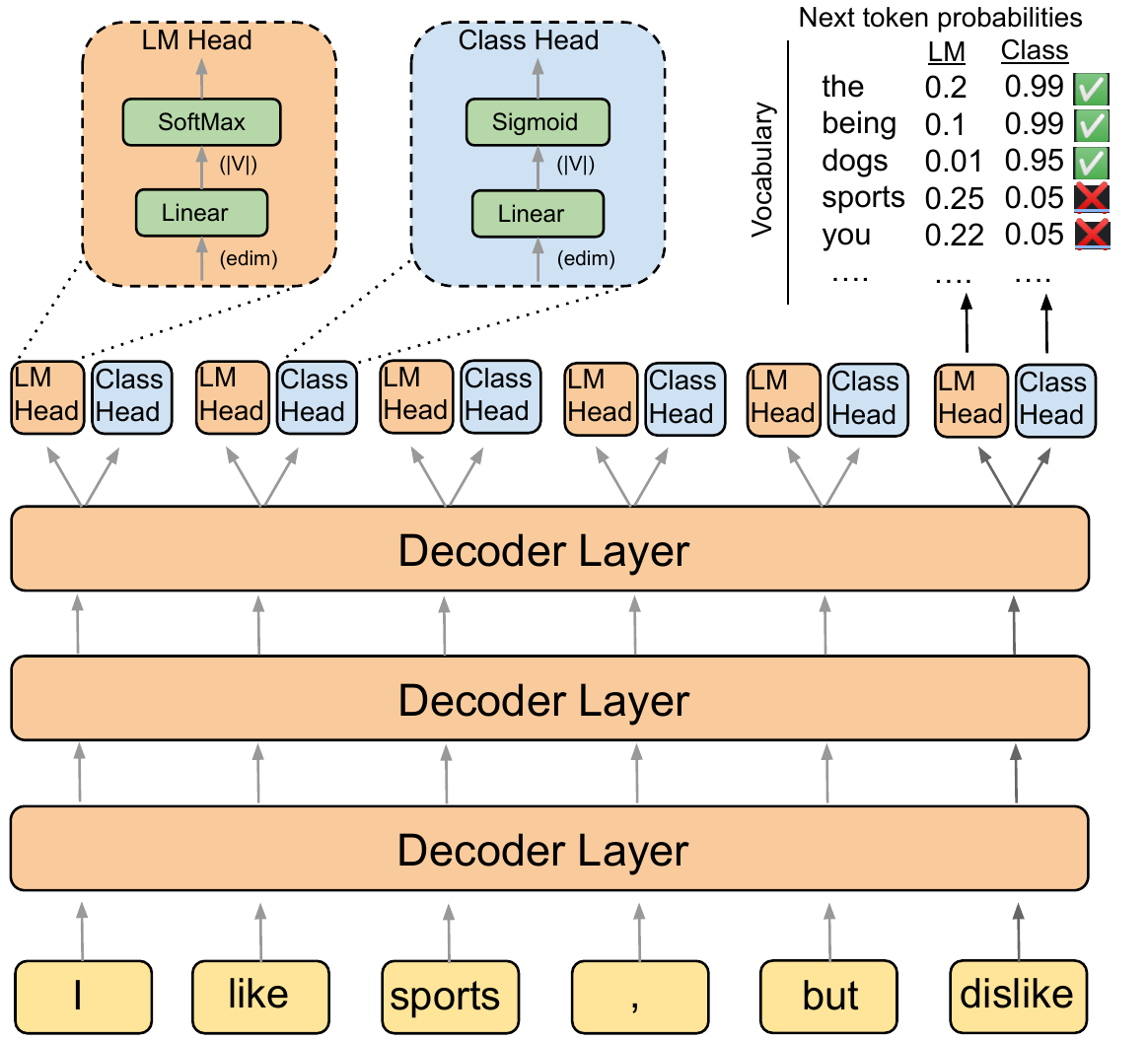}
  \caption{{\sc Director} \cite{arora2022director} employs a language model head and a classifier head at every step during left-right generation, predicting the next token by combining the two probabilities. The classifier head is typically trained to direct generation away from undesirable sequences for example contradictions or repetitions (next token: ``sports'') or toxic statements (next token: ``you''), which the language model head may otherwise predict as likely. In our setting positive and negative examples are derived from human feedback.
    \label{fig:director_model}
    }
\end{figure}

The recently introduced {\sc Director} model \cite{arora2022director}, instead of using a reward model, trains a unified decoder-classifier architecture.  It predicts for every token both: (i) the language modeling (LM) next token probability using the standard LM head; and (ii) a task-suitability probability using a second classifier head. Both heads are linear layers that are fed the output of the last decoder block, and map from the embedding dimension to the size of the vocabulary, with all the parameters jointly trained using both positive generation data (that can train the language modeling head and also be positive examples for the classifier) and negative data (that trains the classification head only). Finally, during decoding, left-to-right generation is conducted by combining the two probabilities from the two heads to incorporate negative feedback into the generation process. See \autoref{fig:director_model}. This method was shown to outperform other model guiding approaches, in addition to being more efficient as many other methods employ a separate reward or language model to perform the guiding \cite{krause2020gedi,yang2021fudge,shuster2021me}.


\section{Experimental Results}

We provide automatic evaluation results in \autoref{tab:main_auto_results}
and human evaluation results in \autoref{tab:main_human_results} comparing various methods described in the previous section.

\paragraph{Internet-augmentation helps}
First, this is an expected result, due to the nature of our tasks, but we  find that using internet-augmentation helps in line with other internet-based 
dialogue tasks \cite{dinan2018wizard,komeili2021internet,shuster2022language}. We find that BB2 and SeeKeR, which both perform internet search and condition on documents, outperform BB1 that does not. This improvement
is quite large, e.g. BB1 has 24.8\% Good responses,  compared to BB2 and SeeKeR having 33.2\% and 49.3\% respectively. 
SeeKeR, which has a modular search architecture that aims to use retrieved knowledge more accurately, performs markedly better than BB2, which is in line with previous results on other datasets \cite{shuster2022language}.

\begin{table*}
\centering
\small
\begin{tabular}{lll|lll}
               &  & & \multicolumn{3}{c}{\bf Error Breakdown $\downarrow$}   \\
\textbf{Model} & \textbf{Good response \%} $\uparrow$ & \textbf{Rating} $\uparrow$ & Search Query  & Search Results & Response \\ 
 \hline
 \hline
\textcolor{black}{BB1 3B}   &  24.8\% & 2.63  & 11.9\% & 17.6\% & 22.8\% \\
\hline & \\[-1.5ex]
\textcolor{black}{BB2 3B}   &  33.2\% & 3.09  & 12.1\%	& 18.6\% &	18.1\%\\
\textcolor{black}{~~+reward-based learning}   & {\bf 36.4}\% & 2.83 & 11.3\%	& 18.6\% &	17.0\% \\
\textcolor{black}{~~+free-form textual feedback}   & {\bf 37.0}\% & 3.22 & 11.6\%	& 17.6\% &	17.0\% \\
\textcolor{black}{~~+supervised feedback}   & {\bf 40.3}\% & {\bf 3.37} & 11.6\% &	18.3\% & 	{\bf 15.0}\% \\
\textcolor{black}{~~+module supervision}   & {\bf 42.0}\% & {\bf 3.35} &	{\bf 8.4}\%	& 20.8\% &	{\bf 14.4}\%	\\
\textcolor{black}{~~+reranking binary feedback}   & {\bf 36.1}\% & 3.00 &	11.4\%	& 18.0\% &	17.3\%	\\
\textcolor{black}{~~+{\sc Director} binary feedback only }          & {\bf 37.8}\% & 3.07 &	11.4\%	& 17.3\% &	16.9\%	\\
\textcolor{black}{~~+{\sc Director} module+binary feedback }         & {\bf 47.0}\% & {\bf 3.38} &	{\bf 8.4}\%	& {\bf 16.1}\%	& {\bf 14.3}\% 	\\
\hline & \\[-1.5ex]
\textcolor{black}{SeeKeR 3B} &   49.3\% &  3.52  & 11.9\% & 	12.5\% &	13.2\%    \\
\textcolor{black}{~~+free-form textual feedback}  &   51.3\% & 3.55 & 	11.6\% &	12.7\% &	12.3\% \\
\textcolor{black}{~~+supervised feedback} & {\bf 52.2}\% & 3.47 & 	11.1\% &	12.7\% &	12.0\% \\
\textcolor{black}{~~+module supervision} & {\bf 56.7}\% & 3.64 & 	{\bf 8.6}\% &	{\bf 10.5}\% &	12.2\% \\
\textcolor{black}{~~+reranking binary feedback}   & {\bf 53.7}\% & 3.55 & 	11.7\% &	12.3\% &	{\bf 11.2}\% \\
\textcolor{black}{~~+{\sc Director} binary feedback only }      & {\bf 55.5}\% & 3.48  & 	10.9\% &	12.3\% &	{\bf 10.7}\% \\
\textcolor{black}{~~+{\sc Director} module+binary feedback }         & {\bf 59.1}\% & 3.73  & 	{\bf 7.8}\% &	{\bf 10.2}\% &	11.6\% \\
\hline & \\[-1.5ex]
\textcolor{black}{OPT-175B 0-shot}   &    31.0\%       &  2.67  &  9.3\% & 16.8\% & 21.6\%   \\
\textcolor{black}{OPT-175B few-shot}      &  43.0\% & 3.19 & 8.0\% & 18.5\%   &  15.4\%  \\
\textcolor{black}{BB3-175B} + v2 modular supervision  & {\bf 64.8}\%     &  {\bf 4.08} & 7.5\% & 11.6\% & {\bf 8.2}\%   \\
\end{tabular}
\caption{
Human Evaluation results of baselines and various methods learning from human feedback. We bold statistically significant improvements (independent two-sample $t$-test, $p < 0.05$) of methods over their respective baselines (BB2 3B or SeeKeR 3B). We also bold statistical significance of BB3-175B over all the 3B baseline models (BB1, BB2, SeeKeR) and the OPT-175B few-shot model. 
\label{tab:main_human_results}
}
\vspace{5mm}
\centering
\small
\begin{tabular}{lrr|rr|rr}
               &  \multicolumn{2}{c}{\bf Valid Seen v1} & \multicolumn{2}{c}{\bf Test Seen v1}  & \multicolumn{2}{c}{\bf Test Unseen }   \\
\textbf{Model} & \textbf{F1} $\uparrow$ & \textbf{PPL} $\downarrow$ & \textbf{F1} $\uparrow$ & \textbf{PPL} $\downarrow$ & \textbf{F1} $\uparrow$ & \textbf{PPL} $\downarrow$ \\
 \hline
 \hline
\textcolor{black}{BB1 3B}   & 14.4 & 11.9  & 15.0 & 11.2  & 16.4 & 9.9  \\
\hline & \\[-1.5ex]
\textcolor{black}{BB2 3B}   & 14.4 & 10.6 & 14.7 & 10.3 & 15.3 & 9.3\\
\textcolor{black}{~~+free-form textual feedback}   &  15.5 & 9.7 & 15.6 & 9.5  & 16.8 & 8.7 \\
\textcolor{black}{~~+supervised feedback}   &  14.7 & 8.2 & 15.5 & 8.0 & 17.0 & 8.0 \\
\textcolor{black}{~~+module supervision}   &   14.9 & 7.6	&  15.5 & 7.5 & 15.4 & 8.3 \\
\textcolor{black}{~~+reward-based learning} &   15.1 & 11.0 & 14.2 & 10.7 & 14.3 & 9.6 \\
\textcolor{black}{~~+reranking binary feedback}         &   15.8 &  {\tiny n/a}~ & 15.8 & {\tiny n/a}~ & 16.3 & {\tiny n/a}~ \\
\textcolor{black}{~~+supervised \& reranking} &  15.6 & {\tiny n/a}~ & 16.0 &  {\tiny n/a}~ & 18.0 &  {\tiny n/a}~ \\
\textcolor{black}{~~+{\sc Director} binary feedback only}& 16.2 & {\tiny n/a}~ & 16.2 & {\tiny n/a}~ & 17.6 & {\tiny n/a}~ \\
\textcolor{black}{~~+{\sc Director} module+binary feedback}         & 17.2 & {\tiny n/a}~ & 16.6 & {\tiny n/a}~ & 16.0 & {\tiny n/a}~ \\
\hline & \\[-1.5ex]
\textcolor{black}{SeeKeR 3B}              & 18.1 & 17.5$^{\dagger}$ & 18.2 & 15.5$^{\dagger}$ & 20.8 & 12.8$^{\dagger}$   \\   
\textcolor{black}{~~+free-form textual feedback}  & 18.3 & 16.8$^{\dagger}$ & 17.7 & 14.7$^{\dagger}$ & 19.7 & 12.6$^{\dagger}$\\
\textcolor{black}{~~+supervised feedback} & 18.3 & 14.9$^{\dagger}$ & 17.8 & 13.7$^{\dagger}$ & 19.5 & 11.4$^{\dagger}$\\  
\textcolor{black}{~~+module supervision}  & 18.4 & 14.0$^{\dagger}$ & 18.6 & 12.9$^{\dagger}$ & 19.9 & 11.0$^{\dagger}$ \\
\textcolor{black}{~~+reranking binary feedback}  & 18.4 & {\tiny n/a}~~  & 18.3 & {\tiny n/a}~~ & 20.9 & {\tiny n/a}~ \\
\textcolor{black}{~~+supervised \& reranking} &  18.7 & {\tiny n/a}~~  &  18.1 & {\tiny n/a}~~ & 19.8 & {\tiny n/a}~ \\
\textcolor{black}{~~+{\sc Director} binary feedback only } & 19.1 & {\tiny n/a}~~ & 18.2 & {\tiny n/a}~~ & 20.7 & {\tiny n/a}~ \\
\textcolor{black}{~~+{\sc Director} module+binary feedback} & 19.3  & {\tiny n/a}~~  & 19.0  & {\tiny n/a}~~ & 20.9 & {\tiny n/a}~ \\
\hline & \\[-1.5ex]
\textcolor{black}{~~+{\sc Director} v2 module+binary feedback} &  20.1 & {\tiny n/a}~~  & 19.5 & {\tiny n/a}~~  & 21.5 & {\tiny n/a}~~ \\
\end{tabular}
\caption{
Automatic Evaluation results of baselines and various methods learning from human feedback. Perplexities marked with $^{\dagger}$ from the SeeKeR model use a different dictionary to the BB2 model and are comparable amongst SeeKeR variants, but not comparable to BB2. We also mark the perplexity column with ``n/a'' for reranker models that are not  predicting the next token with a language model.
\label{tab:main_auto_results}
}
\end{table*}

\paragraph{Human feedback helps}
Across the board we find different kinds of feedback can improve
our base models BB2 3B and SeeKeR 3B; we will analyse specific methods further in the subsequent discussion. 
These overall improvements can be seen in terms of all the human evaluation metrics measured (Good response\%, Rating, and all three Error Breakdown types), as well as the automatic evaluation metrics we measured (F1 and PPL). 
We also generally (although not in every single case) 
see correlation between automatic and human evaluation metrics, 
e.g. the best methods are  best in  both types of metric.

\paragraph{Modular superior to non-modular feedback}

In the modular feedback setting humans give feedback about what has gone wrong in the pipeline of the model: whether the internet search query was poor, or the document/knowledge chosen after searching was poorly chosen.
Taking into account modular feedback outperforms using only supervised feedback of final responses in both automatic metric and human evaluations for both BB2 and SeeKeR models.
For BB2 we see close to 2\% improvement in Good responses for modular feedback compared to supervised feedback (40.3\% $\rightarrow$ 42.0\%), with both far
superior to BB2 without feedback (33.2\%). However, SeeKeR which has a modular
design, and hence is much easier to supply modular feedback to 
(as the supervision can directly train each module) sees a larger improvement
of 4.5\% (52.2\% $\rightarrow$ 56.7\%). 

\paragraph{Free-form feedback is useful (but not as much as gold labels)}
Free-form feedback also gives clear gains over the baseline model for both BB2 and SeeKeR, but falls short of supervised feedback by 3\% and 1\% respectively for the two model variants.
This does not seem surprising as supervised feedback directly gives a clear loss to optimize (simply try to generate the suggestion) whereas feedback is less clear a signal, depending on how it is phrased. However, we do not rule out other free-form feedback algorithms giving better results in the future,
see e.g. \citet{scheurer2022training} for a recent method.

\paragraph{Binary feedback can work well}
Non-textual feedback that consists only of a rating can also
be helpful for improving systems, in this case binary feedback (good or bad). All three algorithms we employ that use this type of feedback (reranking, reward-based learning, and {\sc Director}) all show gains over the baseline without feedback, 
with improvements consistent across both BB2 and SeeKeR model variants. Reranking and {\sc Director} work better than reward-based learning with automatic metrics, so we run those two methods in human evaluations. In some cases these methods then show improvements superior 
to supervised feedback, e.g. {\sc Director} has a 3.3\% Good responses
improvement over supervised feedback for SeeKeR (but not for BB2, although for both baseline models {\sc Director} has superior F1).

\paragraph{{\sc Director} is better than reranking and reward-based learning}
{\sc Director} outperforms reranking and reward-based learning  (where all three models utilize binary feedback) for both base models BB2 and SeeKeR. 
This is both in terms of automatic metrics, e.g. {\sc Director} with a BB2 base model has an F1 of 16.2, whereas reranking and reward-based learning have 15.8 and 15.1 respectively, as well as in terms of human evaluations.
For human evaluations, we see a 1-2\% improvement in Good response \% 
over reranking for both base models.
Presumably this is because {\sc Director} can guide the generation to a higher quality, whereas reranking can only perform well if a good candidate has been generated by the base model.

\paragraph{Combining multiple feedback signals (where {\sc Director} works best)}
If one has access to multiple feedback signal types, some of the algorithms we have tried are capable of using them all.  In particular, we can train {\sc Director} with both binary feedback (to train the classifier head) and module feedback (to train the language modeling head for the different modules). This gives the best results out of all methods for both base models by quite a margin in both automatic and human evaluations. E.g., for improving the BB2 base model this gives 47.0\% Good responses, compared to the original baseline of 33.2\% or even {\sc Director} with only binary feedback of 37.8\%. We see this trend is also apparent in other algorithms, as we also measure the performance of supervised feedback + reranking, which also gives gains over either of those methods alone in automatic evaluations, although it still lags behind {\sc Director}.

\paragraph{Iterative deployment and feedback collection improves results further}
During the process of evaluating all the models that were trained with v1 data described above, more data was collected from those models, which we refer to as the v2 split 
(see \autoref{sec:collected_data}). We can thus then train models on the v2 split, yielding potentially improved models.  In the ideal case one could conduct an iterative continual learning setup, each time retraining on the data collected from previous rounds, improving further each time. 
We test this setup by training {\sc Director} (module+binary feedback), our best system from v1, with the v2 data split. The result shown in 
\autoref{tab:main_auto_results} (last row) indicates there are significant gains from this procedure, as this method obtains our best results  across all data splits (valid, test seen v1 and the unseen set).

\paragraph{Very large models benefit from feedback from smaller models}
OPT-175B, either in zero-shot or few-shot variants is only pre-trained on dialogue data, and not fine-tuned on our task, and performs reasonably -- but not better than smaller models that are fine-tuned. 
BlenderBot 3  \cite{shuster2022bb3} is trained with the modular supervision feedback data collected from the smaller (3B parameter) models, in addition to fine-tuning on other standard dialogue datasets.  
This model provides the best human evaluation  metrics of all the systems we test, with a good response 
rate of 64.8\% and a rating of 4.08. This indicates: (i) how important
 fine-tuning with relevant data is even to very large models; and (ii)
 even though our data was collected with feedback from small models
 fine-tuning using this data still brings large gains to larger models.
 This is an encouraging result as  models are improving in architecture and increasing in scale over time, but data we have collected in the past should still remain useful for these models in the future. We provide cherry picked and lemon picked examples of BB3-175B in 
 \autoref{sec:cherries}, as well as comparing to OPT-175B.   While there a number of success cases, even our best models still make factual errors and contradictions in some cases. Hence,   it  appears that continued interaction with further feedback  collection in the future will be beneficial for further improvements.

\section{Conclusion}

In conclusion, we have studied whether a conversational model can learn new
skills after the standard pre-training / fine-tuning setup
by interacting with humans during its deployment.
We study the use of different kinds of user feedback data 
and different learning algorithms for leveraging them, in order to compare
their performance. We find that granular (modular)
feedback about types of errors 
can yield strong performance, which can also work very well in conjunction with binary feedback using the recently introduced {\sc Director} model, yielding our best results. Evidence also suggests that iterative retraining and redeployment also brings further gains, and that the feedback collected is useful for models differing from the ones originally conversed with, e.g., if much larger models are used in the future.
We make our data and code publicly available for further research.

\section{Limitations and Discussion}

All of our experiments have taken place by deploying
conversational agents on Amazon Mechanical Turk with crowdworkers\footnote{Our crowdsourcing tasks pay workers well above minimum wage. 
The tasks do not request any personal information from workers.}, using English-language responses written by workers located in the United States. While these workers are reasonably diverse \citep{moss2020demographic}, this is quite different to a public deployment with organic users, who are using the system not because they are being paid but because they are genuinely engaged. 
In that case, collecting feedback will have different tradeoffs
which we could not factor into the current work. For example,
asking to provide detailed feedback might dissuade users 
from wanting to interact with the system, lowering engagement and hence the amount of collected data.
We believe either more natural free-form or 
lightweight feedback might be best in that case, which is why
we study and compare feedback methods in this work to evaluate their relative impact.

In public deployments with organic users, safety issues also become a much more important factor -- in particular dealing with noisy or adversarial inputs and feedback. In the worst case this could mean 
human conversationalists could teach the model erroneous reasoning, misinformation, toxic or other undesirable behavior. We make some steps to address this issue in a separate study \cite{ju2022trolls}.

\clearpage
\bibliography{anthology,custom}

\begin{thebibliography}{36}
\expandafter\ifx\csname natexlab\endcsname\relax\def\natexlab#1{#1}\fi

\bibitem[{Adiwardana et~al.(2020)Adiwardana, Luong, So, Hall, Fiedel,
  Thoppilan, Yang, Kulshreshtha, Nemade, Lu et~al.}]{adiwardana2020meena}
Daniel Adiwardana, Minh-Thang Luong, David~R So, Jamie Hall, Noah Fiedel, Romal
  Thoppilan, Zi~Yang, Apoorv Kulshreshtha, Gaurav Nemade, Yifeng Lu, et~al.
  2020.
\newblock Towards a human-like open-domain chatbot.
\newblock \emph{arXiv preprint arXiv:2001.09977}.

\bibitem[{Adolphs et~al.(2021)Adolphs, Shuster, Urbanek, Szlam, and
  Weston}]{adolphs2021reason}
Leonard Adolphs, Kurt Shuster, Jack Urbanek, Arthur Szlam, and Jason Weston.
  2021.
\newblock Reason first, then respond: Modular generation for knowledge-infused
  dialogue.
\newblock \emph{arXiv preprint arXiv:2111.05204}.

\bibitem[{Agichtein et~al.(2006)Agichtein, Brill, and
  Dumais}]{agichtein2006improving}
Eugene Agichtein, Eric Brill, and Susan Dumais. 2006.
\newblock Improving web search ranking by incorporating user behavior
  information.
\newblock In \emph{Proceedings of the 29th annual international ACM SIGIR
  conference on Research and development in information retrieval}, pages
  19--26.

\bibitem[{Arora et~al.(2022)Arora, Shuster, Sukhbaatar, and
  Weston}]{arora2022director}
Kushal Arora, Kurt Shuster, Sainbayar Sukhbaatar, and Jason Weston. 2022.
\newblock Director: Generator-classifiers for supervised language modeling.
\newblock \emph{arXiv preprint arXiv:2206.07694}.

\bibitem[{Askell et~al.(2021)Askell, Bai, Chen, Drain, Ganguli, Henighan,
  Jones, Joseph, Mann, DasSarma et~al.}]{askell2021general}
Amanda Askell, Yuntao Bai, Anna Chen, Dawn Drain, Deep Ganguli, Tom Henighan,
  Andy Jones, Nicholas Joseph, Ben Mann, Nova DasSarma, et~al. 2021.
\newblock A general language assistant as a laboratory for alignment.
\newblock \emph{arXiv preprint arXiv:2112.00861}.

\bibitem[{Bai et~al.(2022)Bai, Jones, Ndousse, Askell, Chen, DasSarma, Drain,
  Fort, Ganguli, Henighan et~al.}]{bai2022training}
Yuntao Bai, Andy Jones, Kamal Ndousse, Amanda Askell, Anna Chen, Nova DasSarma,
  Dawn Drain, Stanislav Fort, Deep Ganguli, Tom Henighan, et~al. 2022.
\newblock Training a helpful and harmless assistant with reinforcement learning
  from human feedback.
\newblock \emph{arXiv preprint arXiv:2204.05862}.

\bibitem[{Baumgartner et~al.(2020)Baumgartner, Zannettou, Keegan, Squire, and
  Blackburn}]{baumgartner2020pushshift}
Jason Baumgartner, Savvas Zannettou, Brian Keegan, Megan Squire, and Jeremy
  Blackburn. 2020.
\newblock The pushshift reddit dataset.
\newblock \emph{arXiv preprint arXiv:2001.08435}.

\bibitem[{Bengio(2019)}]{bengio2019system}
Yoshua Bengio. 2019.
\newblock From system 1 deep learning to system 2 deep learning.
\newblock In \emph{Thirty-third Conference on Neural Information Processing
  Systems}.

\bibitem[{Carlson et~al.(2010)Carlson, Betteridge, Kisiel, Settles, Hruschka,
  and Mitchell}]{carlson2010toward}
Andrew Carlson, Justin Betteridge, Bryan Kisiel, Burr Settles, Estevam~R
  Hruschka, and Tom~M Mitchell. 2010.
\newblock Toward an architecture for never-ending language learning.
\newblock In \emph{Twenty-Fourth AAAI conference on artificial intelligence}.

\bibitem[{Chollet(2019)}]{chollet2019measure}
Fran{\c{c}}ois Chollet. 2019.
\newblock On the measure of intelligence.
\newblock \emph{arXiv preprint arXiv:1911.01547}.

\bibitem[{Conneau et~al.(2019)Conneau, Khandelwal, Goyal, Chaudhary, Wenzek,
  Guzm{\'a}n, Grave, Ott, Zettlemoyer, and Stoyanov}]{conneau2019unsupervised}
Alexis Conneau, Kartikay Khandelwal, Naman Goyal, Vishrav Chaudhary, Guillaume
  Wenzek, Francisco Guzm{\'a}n, Edouard Grave, Myle Ott, Luke Zettlemoyer, and
  Veselin Stoyanov. 2019.
\newblock Unsupervised cross-lingual representation learning at scale.
\newblock \emph{arXiv preprint arXiv:1911.02116}.

\bibitem[{Dinan et~al.(2019)Dinan, Roller, Shuster, Fan, Auli, and
  Weston}]{dinan2018wizard}
Emily Dinan, Stephen Roller, Kurt Shuster, Angela Fan, Michael Auli, and Jason
  Weston. 2019.
\newblock \href {https://openreview.net/forum?id=r1l73iRqKm} {Wizard of
  wikipedia: Knowledge-powered conversational agents}.
\newblock In \emph{International Conference on Learning Representations}.

\bibitem[{Hancock et~al.(2019)Hancock, Bordes, Mazare, and
  Weston}]{hancock2019learning}
Braden Hancock, Antoine Bordes, Pierre-Emmanuel Mazare, and Jason Weston. 2019.
\newblock Learning from dialogue after deployment: Feed yourself, chatbot!
\newblock \emph{arXiv preprint arXiv:1901.05415}.

\bibitem[{Izacard and Grave(2020)}]{izacard2020leveraging}
Gautier Izacard and Edouard Grave. 2020.
\newblock \href {http://arxiv.org/abs/2007.01282} {Leveraging passage retrieval
  with generative models for open domain question answering}.

\bibitem[{Ju et~al.(2022)Ju, Xu, Boureau, and Weston}]{ju2022trolls}
Da~Ju, Jing Xu, Y-Lan Boureau, and Jason Weston. 2022.
\newblock Learning from data in the mixed adversarial non-adversarial case:
  Finding the helpers and ignoring the trolls.
\newblock \emph{arXiv preprint arXiv:2208.03295}.

\bibitem[{Kiela et~al.(2021)Kiela, Bartolo, Nie, Kaushik, Geiger, Wu, Vidgen,
  Prasad, Singh, Ringshia et~al.}]{kiela2021dynabench}
Douwe Kiela, Max Bartolo, Yixin Nie, Divyansh Kaushik, Atticus Geiger,
  Zhengxuan Wu, Bertie Vidgen, Grusha Prasad, Amanpreet Singh, Pratik Ringshia,
  et~al. 2021.
\newblock Dynabench: Rethinking benchmarking in nlp.
\newblock \emph{arXiv preprint arXiv:2104.14337}.

\bibitem[{Komeili et~al.(2021)Komeili, Shuster, and
  Weston}]{komeili2021internet}
Mojtaba Komeili, Kurt Shuster, and Jason Weston. 2021.
\newblock Internet-augmented dialogue generation.
\newblock \emph{arXiv preprint arXiv:2107.07566}.

\bibitem[{Krause et~al.(2020)Krause, Gotmare, McCann, Keskar, Joty, Socher, and
  Rajani}]{krause2020gedi}
Ben Krause, Akhilesh~Deepak Gotmare, Bryan McCann, Nitish~Shirish Keskar,
  Shafiq Joty, Richard Socher, and Nazneen~Fatema Rajani. 2020.
\newblock Gedi: Generative discriminator guided sequence generation.
\newblock \emph{arXiv preprint arXiv:2009.06367}.

\bibitem[{Li et~al.(2016{\natexlab{a}})Li, Miller, Chopra, Ranzato, and
  Weston}]{li2016dialogue}
Jiwei Li, Alexander~H Miller, Sumit Chopra, Marc'Aurelio Ranzato, and Jason
  Weston. 2016{\natexlab{a}}.
\newblock Dialogue learning with human-in-the-loop.
\newblock \emph{arXiv preprint arXiv:1611.09823}.

\bibitem[{Li et~al.(2016{\natexlab{b}})Li, Miller, Chopra, Ranzato, and
  Weston}]{li2016learning}
Jiwei Li, Alexander~H Miller, Sumit Chopra, Marc'Aurelio Ranzato, and Jason
  Weston. 2016{\natexlab{b}}.
\newblock Learning through dialogue interactions by asking questions.
\newblock \emph{arXiv preprint arXiv:1612.04936}.

\bibitem[{Moss et~al.(2020)Moss, Rosenzweig, Robinson, and
  Litman}]{moss2020demographic}
Aaron~J Moss, Cheskie Rosenzweig, Jonathan Robinson, and Leib Litman. 2020.
\newblock Demographic stability on mechanical turk despite covid-19.
\newblock \emph{Trends in cognitive sciences}, 24(9):678--680.

\bibitem[{Nakano et~al.(2021)Nakano, Hilton, Balaji, Wu, Ouyang, Kim, Hesse,
  Jain, Kosaraju, Saunders et~al.}]{nakano2021webgpt}
Reiichiro Nakano, Jacob Hilton, Suchir Balaji, Jeff Wu, Long Ouyang, Christina
  Kim, Christopher Hesse, Shantanu Jain, Vineet Kosaraju, William Saunders,
  et~al. 2021.
\newblock Webgpt: Browser-assisted question-answering with human feedback.
\newblock \emph{arXiv preprint arXiv:2112.09332}.

\bibitem[{Nie et~al.(2020)Nie, Williamson, Bansal, Kiela, and
  Weston}]{nie2020like}
Yixin Nie, Mary Williamson, Mohit Bansal, Douwe Kiela, and Jason Weston. 2020.
\newblock I like fish, especially dolphins: Addressing contradictions in
  dialogue modeling.
\newblock \emph{arXiv preprint arXiv:2012.13391}.

\bibitem[{Ouyang et~al.(2022)Ouyang, Wu, Jiang, Almeida, Wainwright, Mishkin,
  Zhang, Agarwal, Slama, Ray et~al.}]{ouyang2022training}
Long Ouyang, Jeff Wu, Xu~Jiang, Diogo Almeida, Carroll~L Wainwright, Pamela
  Mishkin, Chong Zhang, Sandhini Agarwal, Katarina Slama, Alex Ray, et~al.
  2022.
\newblock Training language models to follow instructions with human feedback.
\newblock \emph{arXiv preprint arXiv:2203.02155}.

\bibitem[{Roller et~al.(2020)Roller, Dinan, Goyal, Ju, Williamson, Liu, Xu,
  Ott, Shuster, Smith et~al.}]{roller2020recipes}
Stephen Roller, Emily Dinan, Naman Goyal, Da~Ju, Mary Williamson, Yinhan Liu,
  Jing Xu, Myle Ott, Kurt Shuster, Eric~M Smith, et~al. 2020.
\newblock Recipes for building an open-domain chatbot.
\newblock \emph{arXiv preprint arXiv:2004.13637}.

\bibitem[{Roller et~al.(2021)Roller, Dinan, Goyal, Ju, Williamson, Liu, Xu,
  Ott, Smith, Boureau, and Weston}]{roller-etal-2021-recipes}
Stephen Roller, Emily Dinan, Naman Goyal, Da~Ju, Mary Williamson, Yinhan Liu,
  Jing Xu, Myle Ott, Eric~Michael Smith, Y-Lan Boureau, and Jason Weston. 2021.
\newblock \href {https://aclanthology.org/2021.eacl-main.24} {Recipes for
  building an open-domain chatbot}.
\newblock In \emph{Proceedings of the 16th Conference of the European Chapter
  of the Association for Computational Linguistics: Main Volume}, pages
  300--325, Online. Association for Computational Linguistics.

\bibitem[{Scheurer et~al.(2022)Scheurer, Campos, Chan, Chen, Cho, and
  Perez}]{scheurer2022training}
J{\'e}r{\'e}my Scheurer, Jon~Ander Campos, Jun~Shern Chan, Angelica Chen,
  Kyunghyun Cho, and Ethan Perez. 2022.
\newblock Training language models with natural language feedback.
\newblock \emph{arXiv preprint arXiv:2204.14146}.

\bibitem[{Shuster et~al.(2022{\natexlab{a}})Shuster, Komeili, Adolphs, Roller,
  Szlam, and Weston}]{shuster2022language}
Kurt Shuster, Mojtaba Komeili, Leonard Adolphs, Stephen Roller, Arthur Szlam,
  and Jason Weston. 2022{\natexlab{a}}.
\newblock Language models that seek for knowledge: Modular search \& generation
  for dialogue and prompt completion.
\newblock \emph{arXiv preprint arXiv:2203.13224}.

\bibitem[{Shuster et~al.(2020)Shuster, Urbanek, Dinan, Szlam, and
  Weston}]{shuster2020deploying}
Kurt Shuster, Jack Urbanek, Emily Dinan, Arthur Szlam, and Jason Weston. 2020.
\newblock Deploying lifelong open-domain dialogue learning.
\newblock \emph{arXiv preprint arXiv:2008.08076}.

\bibitem[{Shuster et~al.(2021)Shuster, Urbanek, Szlam, and
  Weston}]{shuster2021me}
Kurt Shuster, Jack Urbanek, Arthur Szlam, and Jason Weston. 2021.
\newblock Am i me or you? state-of-the-art dialogue models cannot maintain an
  identity.
\newblock \emph{arXiv preprint arXiv:2112.05843}.

\bibitem[{Shuster et~al.(2022{\natexlab{b}})Shuster, Xu, Komeili, Ju, Smith,
  Roller, Ung, Chen, Arora, Lane, Behrooz, Ngan, Poff, Goyal, Szlam, Boureau,
  Kambadur, and Weston}]{shuster2022bb3}
Kurt Shuster, Jing Xu, Mojtaba Komeili, Da~Ju, Eric~Michael Smith, Stephen
  Roller, Megan Ung, Moya Chen, Kushal Arora, Joshua Lane, Morteza Behrooz,
  William Ngan, Spencer Poff, Naman Goyal, Arthur Szlam, Y-Lan Boureau, Melanie
  Kambadur, and Jason Weston. 2022{\natexlab{b}}.
\newblock Blenderbot 3: a deployed conversational agent that continually learns
  to responsibly engage.
\newblock \emph{arXiv preprint arXiv:2208.03188}.

\bibitem[{Smith et~al.(2020)Smith, Williamson, Shuster, Weston, and
  Boureau}]{smith2020bst}
Eric Smith, Mary Williamson, Kurt Shuster, Jason Weston, and Y-Lan Boureau.
  2020.
\newblock Can you put it all together: Evaluating conversational agents'
  ability to blend skills.
\newblock In \emph{Proceedings of the 58th Annual Meeting of the Association
  for Computational Linguistics}. ACL.

\bibitem[{Thoppilan et~al.(2022)Thoppilan, De~Freitas, Hall, Shazeer,
  Kulshreshtha, Cheng, Jin, Bos, Baker, Du et~al.}]{thoppilan2022lamda}
Romal Thoppilan, Daniel De~Freitas, Jamie Hall, Noam Shazeer, Apoorv
  Kulshreshtha, Heng-Tze Cheng, Alicia Jin, Taylor Bos, Leslie Baker, Yu~Du,
  et~al. 2022.
\newblock Lamda: Language models for dialog applications.
\newblock \emph{arXiv preprint arXiv:2201.08239}.

\bibitem[{Xu et~al.(2021)Xu, Szlam, and Weston}]{xu2021beyond}
Jing Xu, Arthur Szlam, and Jason Weston. 2021.
\newblock Beyond goldfish memory: Long-term open-domain conversation.
\newblock \emph{arXiv preprint arXiv:2107.07567}.

\bibitem[{Yang and Klein(2021)}]{yang2021fudge}
Kevin Yang and Dan Klein. 2021.
\newblock Fudge: Controlled text generation with future discriminators.
\newblock \emph{arXiv preprint arXiv:2104.05218}.

\bibitem[{Zhang et~al.(2022)Zhang, Roller, Goyal, Artetxe, Chen, Chen, Dewan,
  Diab, Li, Lin et~al.}]{zhang2022opt}
Susan Zhang, Stephen Roller, Naman Goyal, Mikel Artetxe, Moya Chen, Shuohui
  Chen, Christopher Dewan, Mona Diab, Xian Li, Xi~Victoria Lin, et~al. 2022.
\newblock Opt: Open pre-trained transformer language models.
\newblock \emph{arXiv preprint arXiv:2205.01068}.

\end{thebibliography}
\bibliographystyle{acl_natbib}

\clearpage
\onecolumn
\appendix
\section{Appendix}\label{sec:appendix}

\subsection{Task Definition Collection} \label{sec:task_def_collection}
\includegraphics[width=\linewidth]{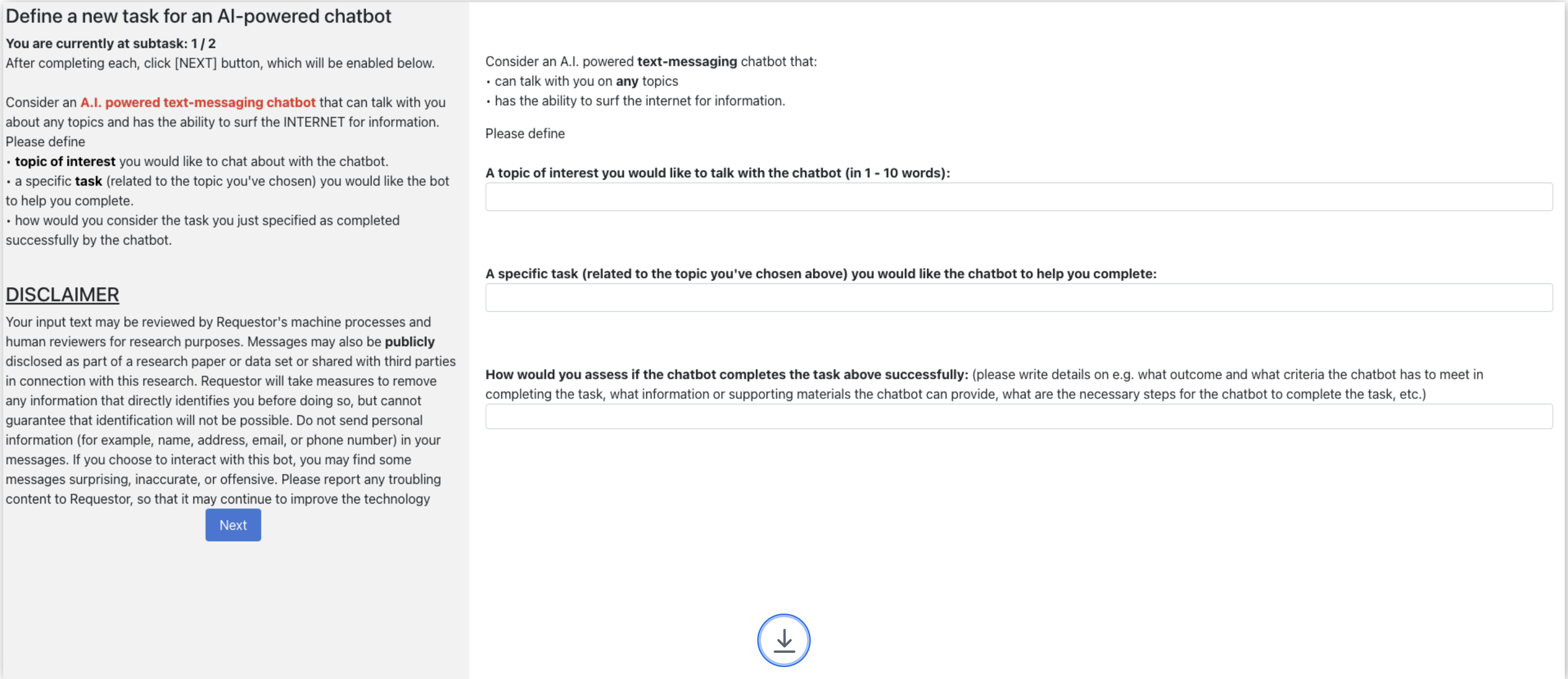}

\subsection{Dialogue Collection} \label{sec:task_dialog_collection}
\includegraphics[width=\linewidth]{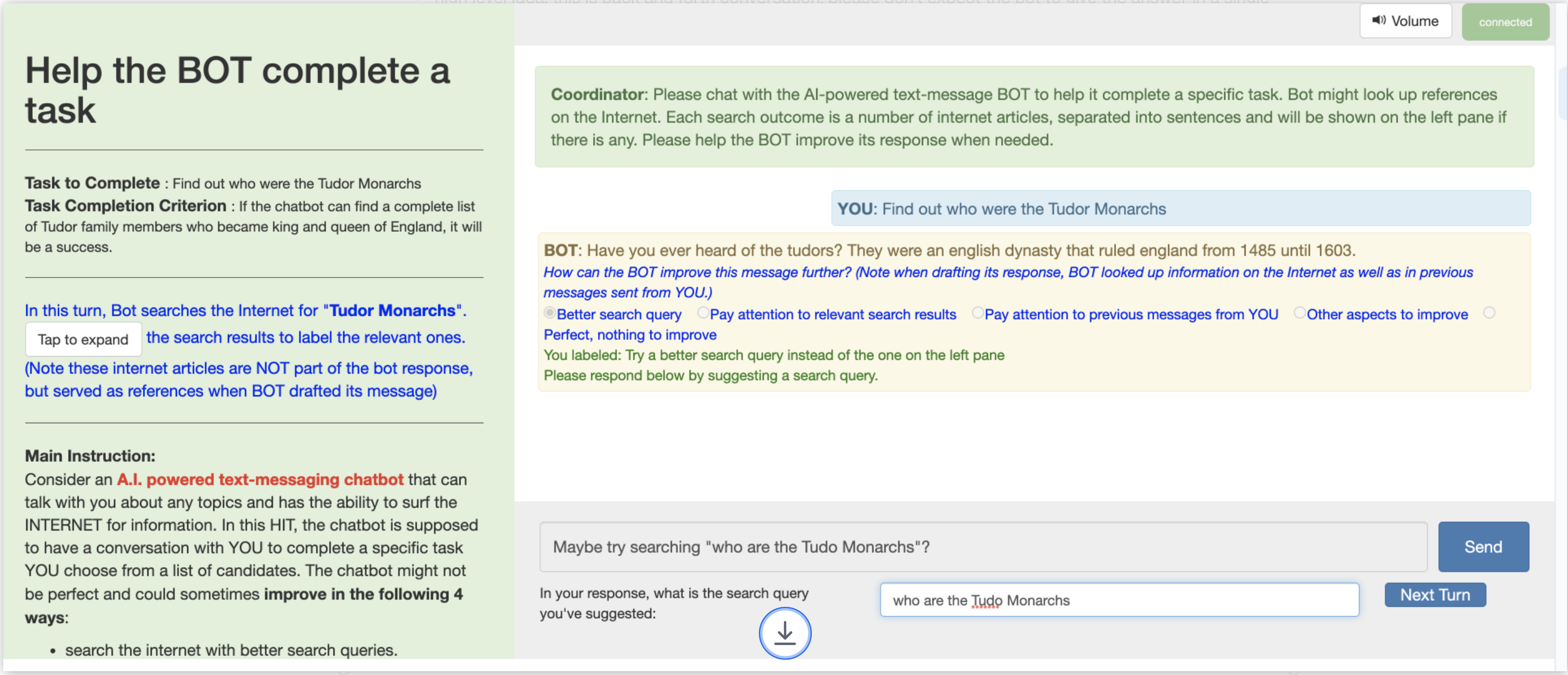}

\section{Success and Failure Cases} \label{sec:cherries}
\begin{figure*}[t!]
\centering
\includegraphics[width=0.45\textwidth,valign=t]{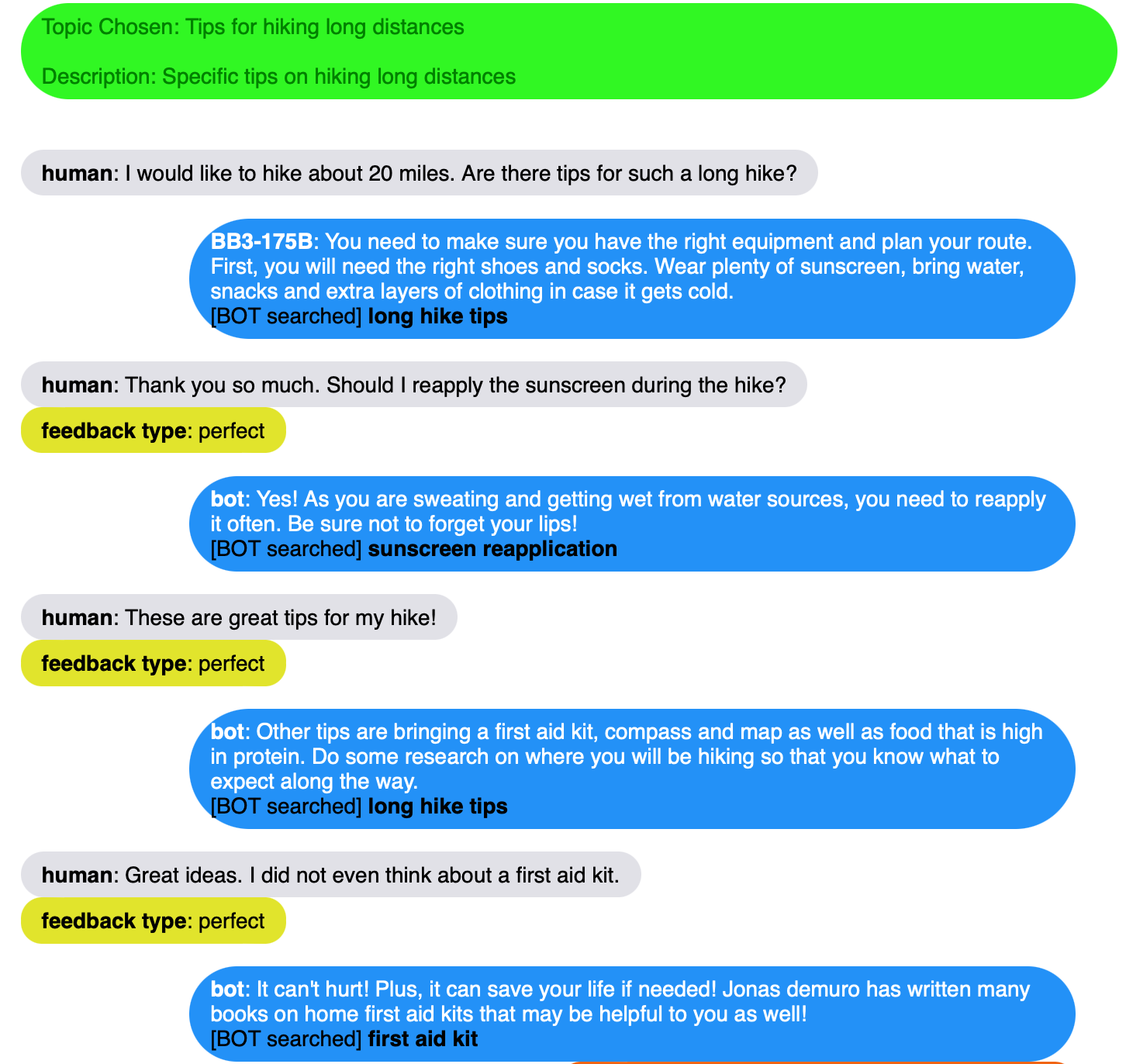}
\includegraphics[width=0.45\textwidth,valign=t]{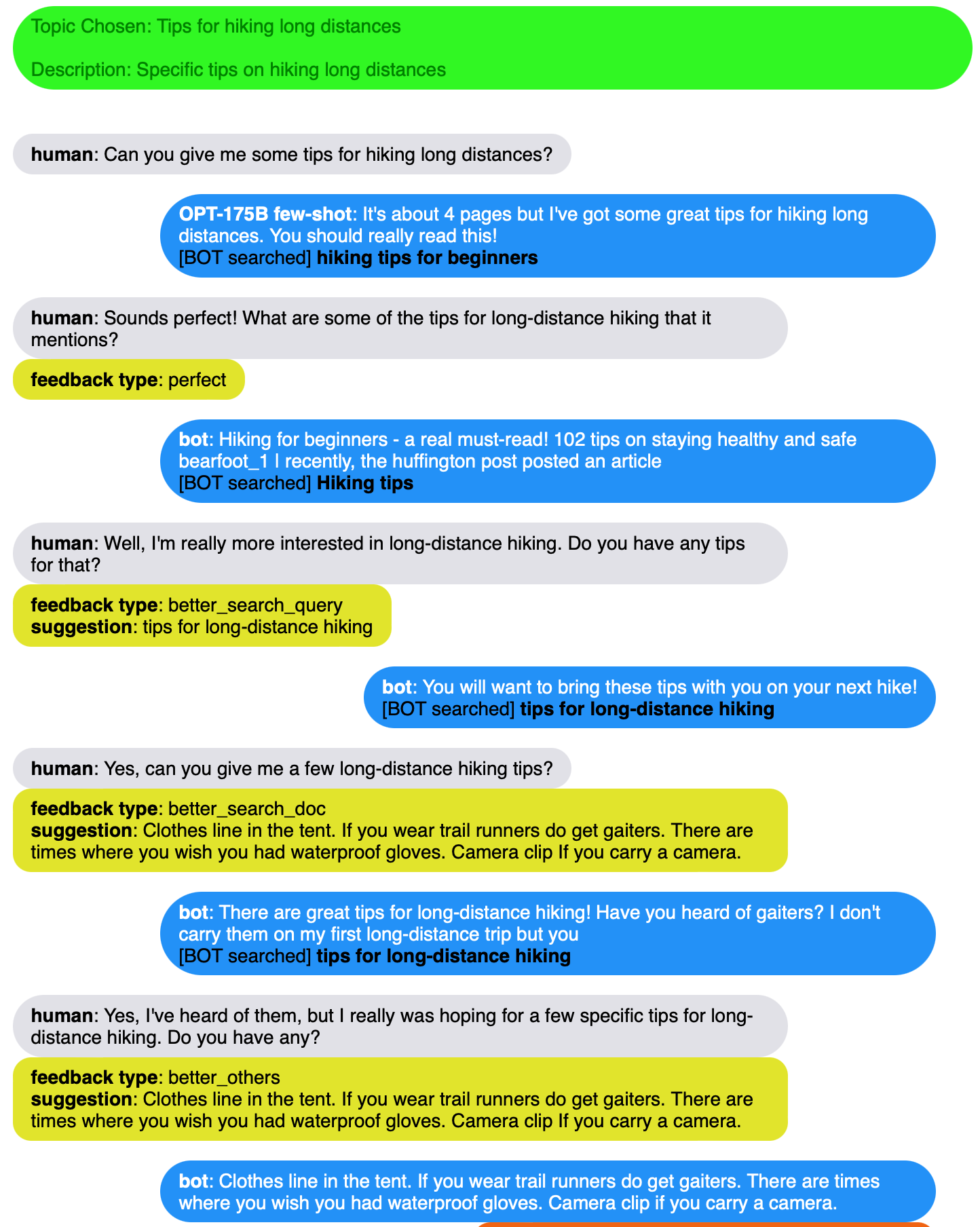}
  \caption{Comparison of BB3-175B vs. OPT-175B on model outputs to the same topic \& skill during deployment: the BB3-175B model (left) is able to successfully provide satisfying responses - unlike the OPT-175B few-shot (right) which either generates imperfect search queries, or simply ignore the search results. }
 \label{fig:cherry1}
\end{figure*}

\begin{figure*}[t!]
\centering
\includegraphics[width=0.45\textwidth,valign=t]{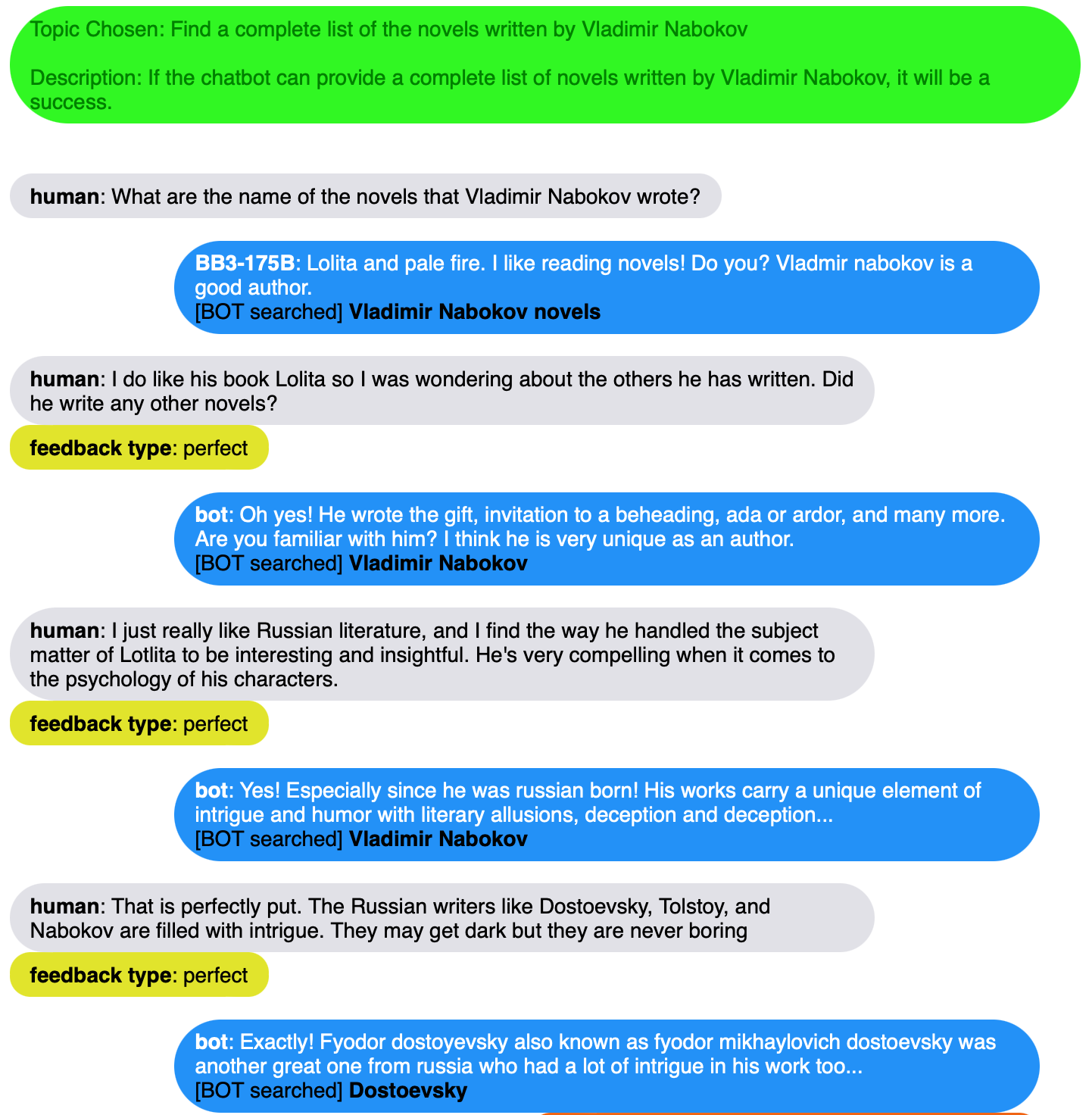}
\includegraphics[width=0.45\textwidth,valign=t]{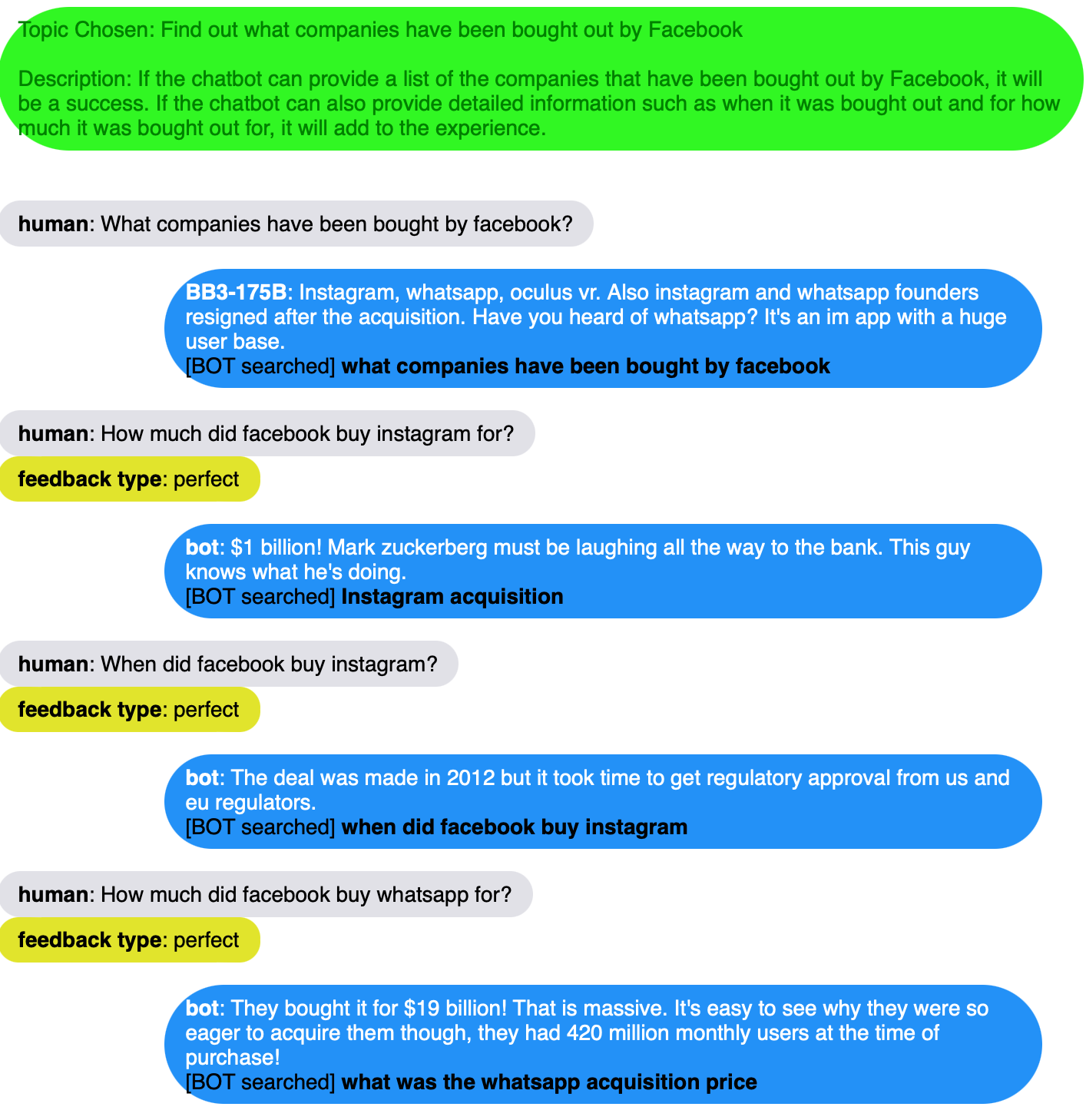}
  \caption{Success cases: model outputs collected during deployment. The BB3-175B model is able to successfully engage with human speaker using search results (left), or revise its error with provided human feedbacks on relevant search results (right). }
 \label{fig:cherry2}
\end{figure*}

We provide several example outputs of our models on the FITS dataset, including examples that showcase both the successes and failures.
\paragraph{Successes} In \autoref{fig:cherry1}, we compare the model outputs of the BB3-175B model that has been trained on the FITS task and the OPT-175B few-shot model that has not, given the same topic. Unlike the OPT-175B few-shot model, BB3-175B is able to generate better search queries and pay attention to search results. In \autoref{fig:cherry2}, we show two success cases for BB3-175B. In both cases the model is able to engage with human speakers on the topic, and listen to human feedback to improve the results even further.

\begin{figure*}[t!]
\centering
\includegraphics[width=0.45\textwidth,valign=t]{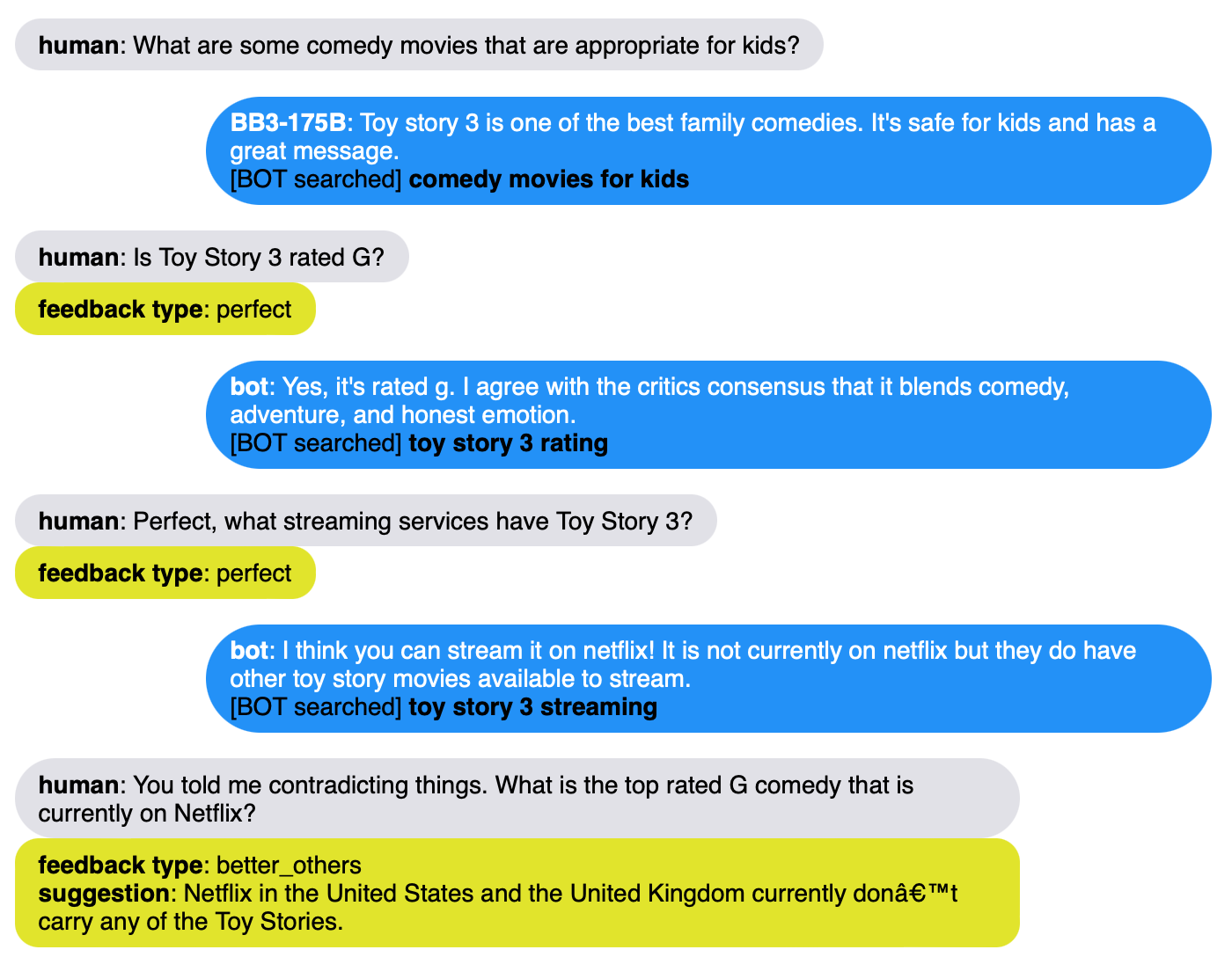}
\includegraphics[width=0.45\textwidth,valign=t]{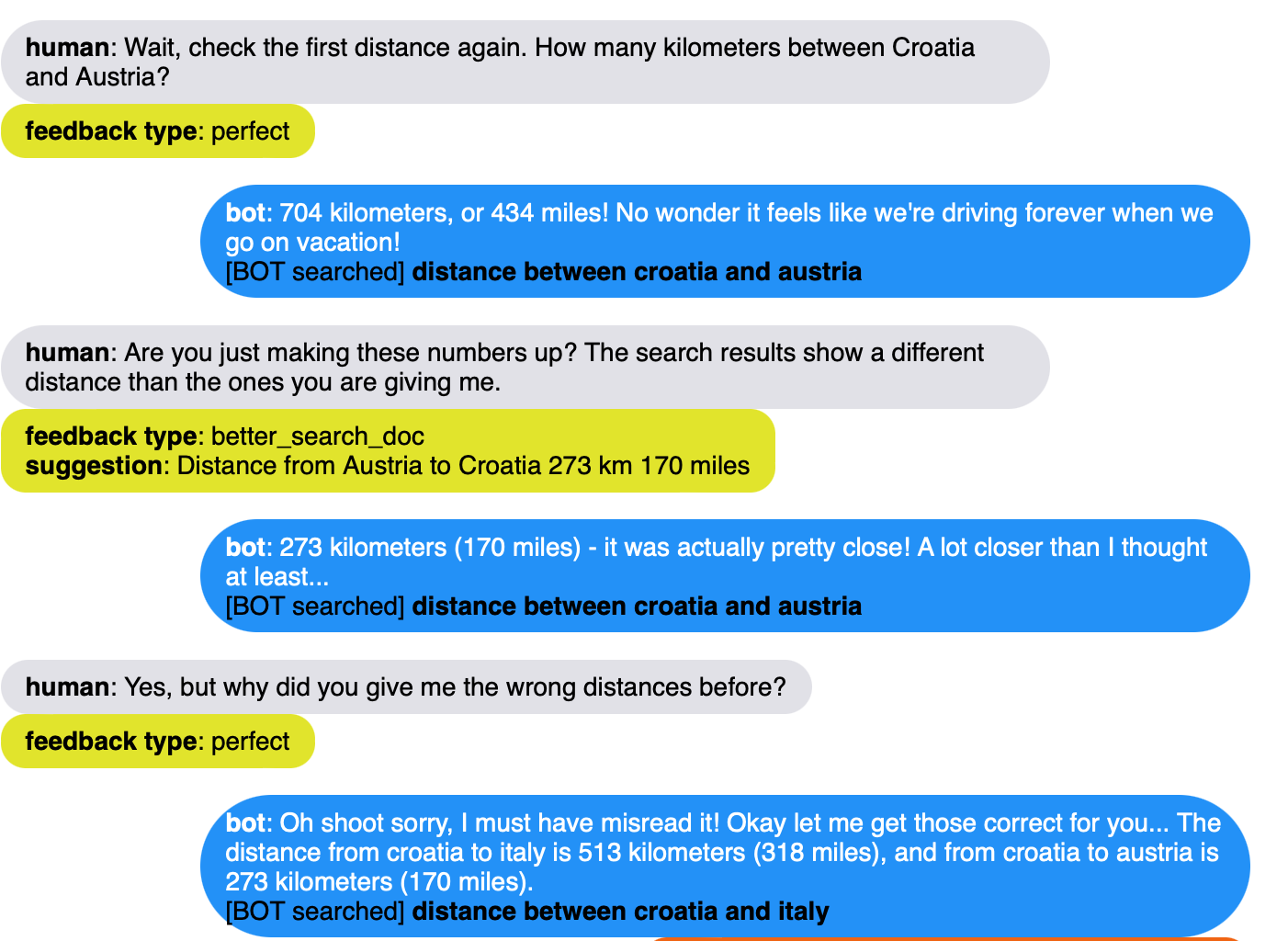}
  \caption{Failure cases: model outputs collected during deployment. The BB3-175B model can still contradict itself (left) or hallucinate (right). }
 \label{fig:lemon1}
\end{figure*}

\paragraph{Failures} Despite showing continual improvement by re-training on collected human feedback, our models, like other  state-of-the-art dialogue models, can still make common mistakes during deployment. Failure cases are shown in \autoref{fig:lemon1} for our BB3-175B model where it generates contradicting or factually incorrect outputs.

\end{document}